\documentclass[letterpaper, 10 pt, conference]{ieeeconf}  %
\usepackage{graphicx}
\usepackage[caption=false,font=scriptsize,labelfont=rm,textfont=rm,subrefformat=parens]{subfig}
\usepackage{amsmath}
\usepackage{multirow}
\usepackage{amssymb}

\IEEEoverridecommandlockouts                              %

\title{\LARGE \bf
Leveraging Temporal Contexts to Enhance Vehicle-Infrastructure Cooperative Perception
}

\author{Jiaru Zhong$^{1,2}$, Haibao Yu$^{3,4}$, Tianyi Zhu$^{2}$, Jiahui Xu$^{2}$, Wenxian Yang$^{3}$, Zaiqing Nie$^{3}$ and Chao Sun$^{1,2,*}$%
\thanks{This work was funded by Key-Area Research and Development Program of Guangdong Province (2023B0909040001).}%
\thanks{$^{1}$Jiaru Zhong and Chao Sun are with Shenzhen Automotive Research Institute, Beijing Institute of Technology, Shenzhen, China. (e-mail: zhongjiaru@bit.edu.cn).}
\thanks{
$^{2}$Jiaru Zhong, Tianyi Zhu, Jiahui Xu and Chao Sun are with the School of Mechanical Engineering, Beijing Institute of Technology, Beijing, China.}
\thanks{
$^{3}$Haibao Yu, Wenxian Yang and Zaiqing Nie are with the Institute for AI Industry Research, Tsinghua University, Beijing, China.}
\thanks{$^{4}$Haibao Yu is also with the School of Computer Science, The University of Hong Kong, Hong Kong, China.}
\thanks{$*$ Corresponding author: Chao Sun. (e-mail: chaosun@bit.edu.cn).}
}

\begin{document}

\maketitle
\thispagestyle{empty}
\pagestyle{empty}

\begin{abstract}
Infrastructure sensors installed at elevated positions offer a broader perception range and encounter fewer occlusions. Integrating both infrastructure and ego-vehicle data through V2X communication, known as vehicle-infrastructure cooperation, has shown considerable advantages in enhancing perception capabilities and addressing corner cases encountered in single-vehicle autonomous driving.
However, cooperative perception still faces numerous challenges, including limited communication bandwidth and practical communication interruptions.
In this paper, we propose CTCE, a novel framework for cooperative 3D object detection. This framework transmits queries with temporal contexts enhancement, effectively balancing transmission efficiency and performance to accommodate real-world communication conditions.
Additionally, we propose a temporal-guided fusion module to further improve performance. The roadside temporal enhancement and vehicle-side spatial-temporal fusion together constitute a multi-level temporal contexts integration mechanism, fully leveraging temporal information to enhance performance.
Furthermore, a motion-aware reconstruction module is introduced to recover lost roadside queries due to communication interruptions.
Experimental results on V2X-Seq and V2X-Sim datasets demonstrate that CTCE outperforms the baseline QUEST, achieving improvements of 3.8\% and 1.3\% in mAP, respectively. Experiments under communication interruption conditions validate CTCE's robustness to communication interruptions.
\end{abstract}

\section{INTRODUCTION}
In recent years, significant progress has been made in autonomous driving, driven by advancements in artificial intelligence. However, single-vehicle autonomous driving still grapples with serious safety challenges arising from various corner cases, such as long-range perception limitations and occlusion issues, which are attributed to the exclusive reliance on ego-vehicle sensor data and its restricted perception range \cite{caillot2022survey}.
Vehicle-infrastructure cooperation, where connected autonomous vehicles (CAVs) leverage both ego-vehicle and infrastructure sensor data through vehicle-to-everything (V2X) communication \cite{han2023collaborative}, holds immense potential for expanding CAVs' perception range and augmenting their driving capabilities. 
3D object detection is a pivotal perception task in autonomous driving, with cameras emerging as low-cost sensors that are widely employed in CAVs and intelligent transportation systems (ITS). Therefore, this paper aims to address the camera-based vehicle-infrastructure cooperative 3D (VIC3D) object detection task.

\begin{figure}[!t]
    \centering
    \subfloat[Single-Frame Spatial Cooperation]{
        \includegraphics[width=0.95\linewidth]{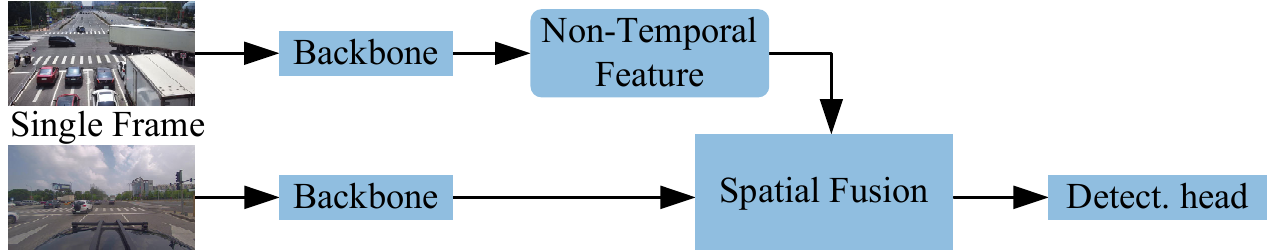} \label{fig1a}}\\
    \vspace{-0.25cm}
    \subfloat[Multi-Frame Spatial-Temporal Cooperation]{
        \includegraphics[width=0.95\linewidth]{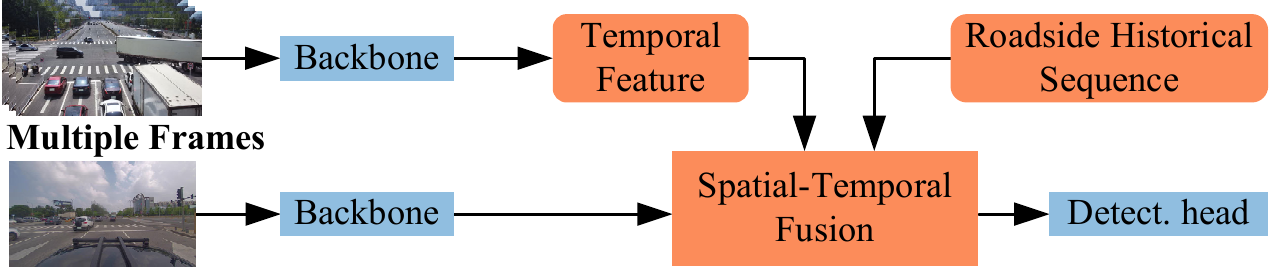} \label{fig1b}}
    \caption{\textbf{The Comparison of Different Cooperation Methods.} In contrast to single-frame spatial cooperation, our proposed multi-frame spatial-temporal cooperation utilizes temporal contexts in two aspects: extracting temporal features from roadside multiple frames and performing spatial-temporal fusion with roadside historical sequence.}
    \label{fig1}
\end{figure}

VIC3D object detection involves the ego-vehicle (EV) aggregating its data and data transmitted by the roadside unit (RSU) to predict the positions, dimensions, and orientation of surrounding objects under limited communication bandwidth.
To leverage roadside data for improving detection performance, there are two critical issues: what to transmit to meet the limited communication conditions, and how to fuse the received data for cooperative detection.
\textbf{For the former issue (what to transmit)}, we choose to transmit queries \cite{fan2023quest}, which are a type of intermediate feature. Compared to transmitting raw images, which entails significant transmission costs, and transmitting detection results, that leads to the loss of crucial information, intermediate features strike a balance in transmission efficiency while retaining essential information \cite{wang2020v2vnet}. Unlike traditional intermediate features, such as BEV features \cite{xu2022v2x}, queries are more sparse at the spatial level and more suitable for transmission. 
Furthermore, as shown in Fig. \ref{fig1}\subref{fig1b}, different from previous works that transmit non-temporal features, we extract temporal information from historical frames and transmit \textit{temporal features}.
\textbf{For the latter issue (how to fuse)}, as depicted in Fig. \ref{fig1}\subref{fig1a}, most previous works typically fuse data from both sides frame-by-frame using a spatial fusion approach. However, this fusion fails to fully exploit the temporal information. In this work, we introduce the temporal fusion with \textit{roadside historical sequence} based on spatial fusion, denoted as \textit{spatial-temporal fusion}, thereby enhancing cooperative detection.

\begin{figure}[!t]
    \centering
    \subfloat[\scriptsize Without Communication Interruption]{    		\includegraphics[width=0.479\linewidth]{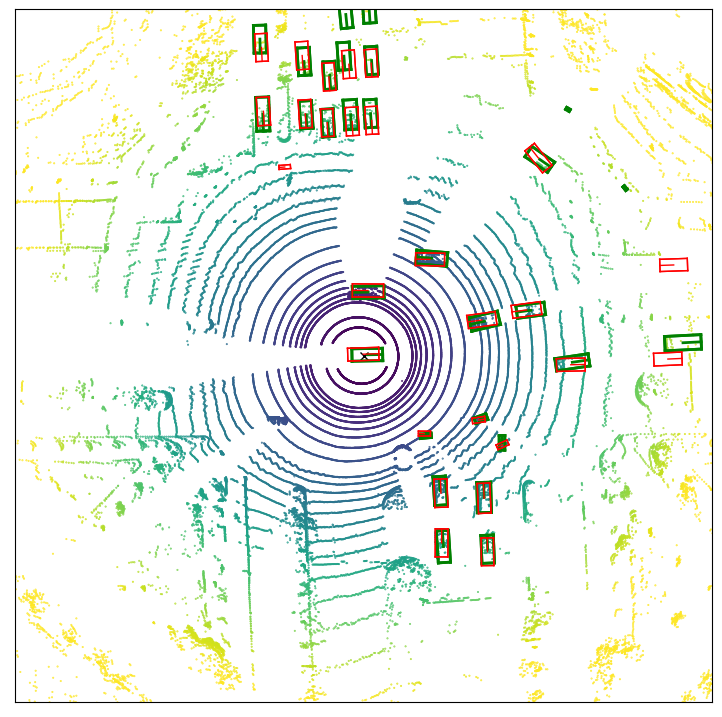}} \label{fig2a}
    \subfloat[\scriptsize With Communication Interruption]{	\includegraphics[width=0.479\linewidth]{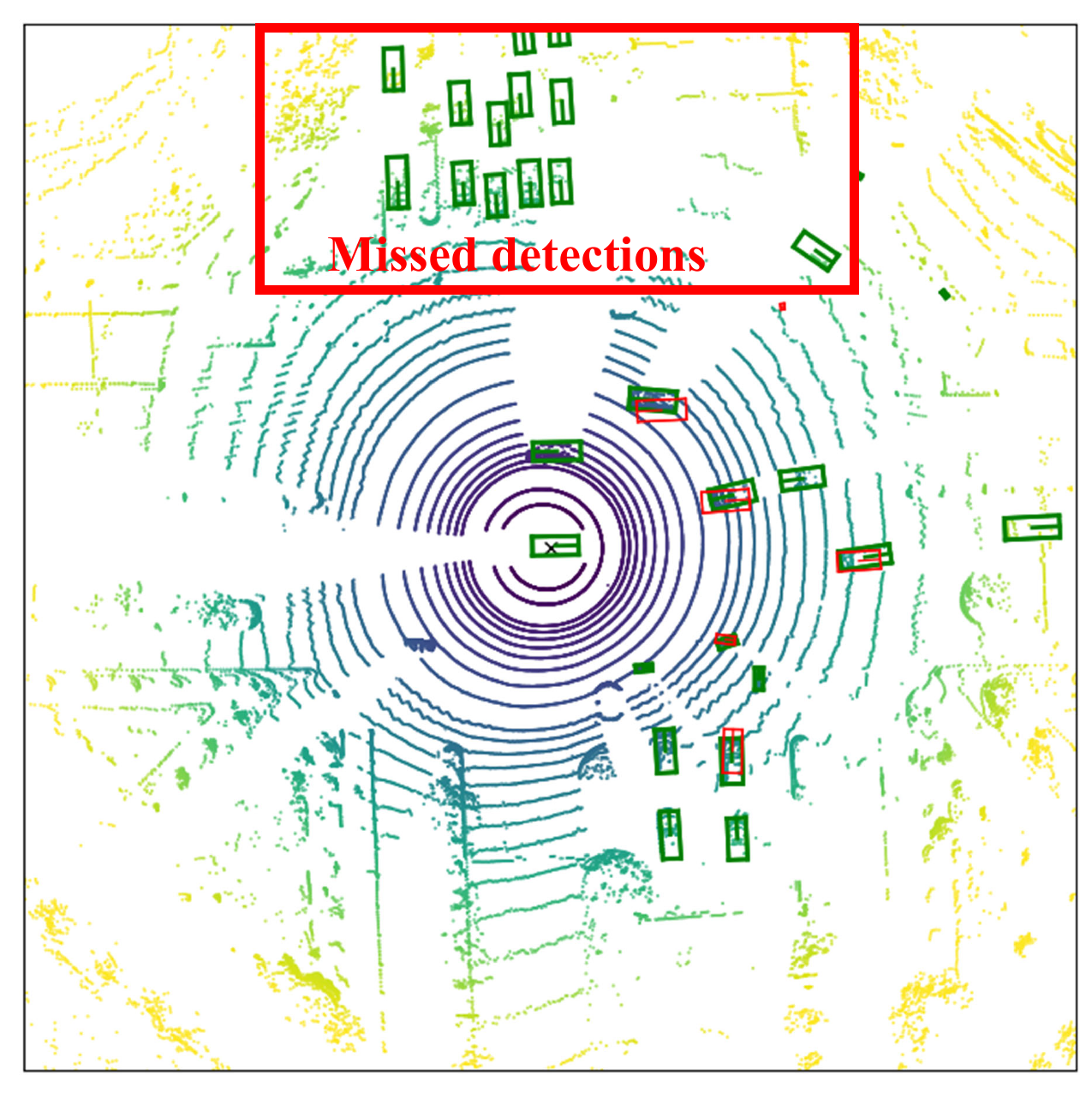}} \label{fig2b}
    \caption{\textbf{The Illustration of Communication Interruption.}
    Red and green boxes denote detection and ground truth results respectively. Compared to ideal communication, interruption can cause transmission loss, which harms cooperative detection.
    }
    \label{fig2}
\end{figure}

In this paper, we propose a Camera-based VIC3D object detection framework with Temporal Contexts Enhancement, dubbed CTCE, depicted in Fig. \ref{fig3}. The primary goal of CTCE is to maximize the utilization of temporal information at multiple levels to enhance cooperative detection. 
At the roadside, we employ a temporal contexts aggregation (TCA) module to consolidate temporal cues from multiple historical frames. 
On the ego-vehicle side, we implement spatial-temporal fusion and utilize a temporal-guided fusion (TGF) module to amalgamate the roadside and ego-vehicle queries, along with the roadside historical sequence. Through such a multi-level temporal contexts integration (MTCI) mechanism, the fused queries exhibit a considerably extended temporal receptive field.
It is worth noting that while recent methods have made efforts to leverage temporal information to solve the latency issue \cite{yu2024flow,dao2024practical,zhang2023robust,yang2024how2comm,wei2024asynchrony} or enhance performance \cite{yang2023spatio,liu2024select2col,wang2023umc,yang2023what2comm,meng2023hydro,hu2024pragmatic}, CTCE stands out by fully capitalizing on the advantages offered by temporal information.
Furthermore, practical communication conditions often entail interruptions \cite{ren2023interruption}. These interruptions can significantly degrade cooperative perception, as illustrated in Fig. \ref{fig2}. 
To enhance robustness, we develop a simple yet effective module, dubbed motion-aware reconstruction (MAR), which directly reconstructs the lost queries from the stored roadside historical sequence.

We evaluate our CTCE on the real-world dataset V2X-Seq \cite{yu2023v2x} and the simulated dataset V2X-Sim \cite{li2022v2x}. CTCE significantly surpasses individual perception and outperforms the baseline QUEST \cite{fan2023quest}, with improvements of $3.8\%$ mAP on V2X-Seq and 
$1.3\%$ mAP on V2X-Sim, respectively. Moreover, we compare our approach with two typical temporal fusion strategies: interacting with fused queries \cite{liu2024select2col} and ego-vehicle queries \cite{yang2023spatio}. The results demonstrate that interacting with roadside historical queries yields greater temporal benefits.
To assess the robustness against communication interruptions, we adopt the packet drop rate (PDR) as a metric to quantify the probability of interruptions. CTCE consistently maintains superior performance compared to other approaches across all PDRs.

Overall, the main contributions can be summarized as:
\begin{itemize}
    \item We propose CTCE, a novel vehicle-infrastructure cooperative 3D object detection framework. To our knowledge, this is the first camera-based temporal cooperative perception framework.
    \item We propose a multi-level temporal contexts integration mechanism, where temporal features are generated at the roadside, and roadside historical sequences are aggregated within spatial-temporal fusion. Additionally, we design a motion-aware reconstruction module to recover lost features during communication interruptions.
    \item Extensive experiments demonstrate that our proposed temporal incorporation modules effectively enhance the performance and that the framework exhibits strong robustness against communication interruptions.
\end{itemize}

\section{RELATED WORK}

\subsection{Individual 3D Object Detection}
3D object detection aims to predict the oriented 3D bounding boxes and the categories of objects which serves as the foundation for subsequent tasks such as prediction \cite{xu2023risk}. Individual methods rely solely on observations from sensors mounted on the vehicles \cite{xu2022unf,zhu2024lanemapnet,liu2024bevmamba}. Depending on the sensors, these methods can be roughly categorized into camera-based, LiDAR-based, and multi-modal-based. BEVFormer \cite{li2022bevformer} and StreamPETR \cite{wang2023exploring} detect objects from surround-view images by leveraging temporal information. Approaches relying on LiDARs, like VoxelNet \cite{zhou2018voxelnet} and PointPillars \cite{lang2019pointpillars}, quantize the point clouds into voxels or pillars and output a birds-eye-view feature map. Multi-modal methods utilize data from heterogeneous sensors such as cameras and LiDARs \cite{wang2023unitr}. Our proposed method differs from the abovementioned methods by leveraging data from infrastructure and vehicle cameras to address the constraints inherent in individual 3D object detection \cite{caillot2022survey, han2023collaborative}.

\begin{figure*}[t]
    \centering
    \includegraphics[width=0.90\linewidth]{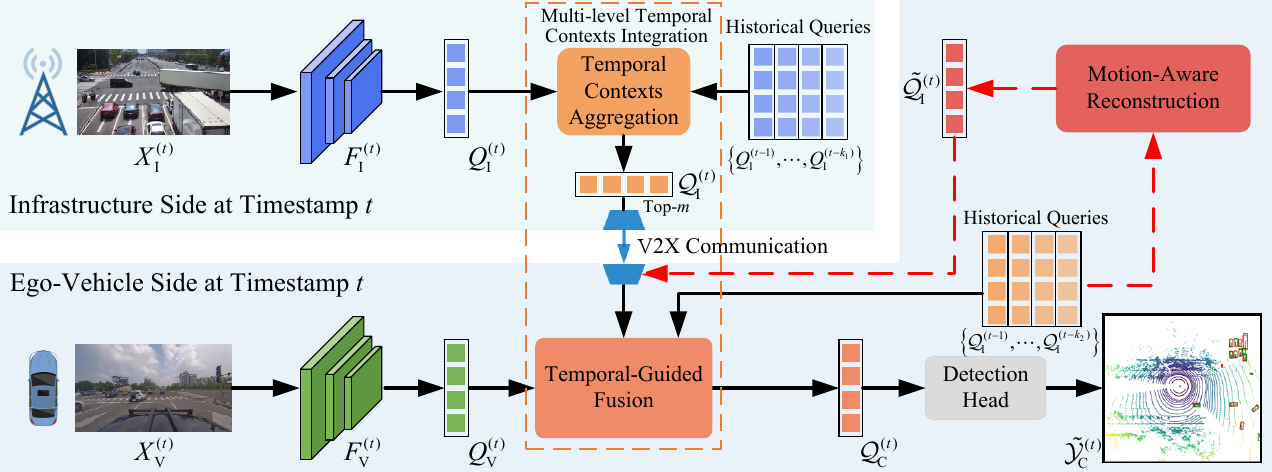}
    \vspace{-0.2cm}
    \caption{\textbf{Overview of the Proposed CTCE.}
    \textbf{At infrastructure side} i) The roadside queries are generated from the image and interact with historical queries to obtain temporal queries. ii) The temporal queries are filtered by confidence and transmitted to the ego-vehicle through V2X communication. \textbf{At ego-vehicle side} i) ego-vehicle queries are extracted from ego-vehicle image. ii) A novel spatial-temporal fusion module fuses ego-vehicle queries, roadside queries, and stored roadside historical queries. iii) The fused queries are input to the detection head to generate the cooperative perception results. iv) a motion-aware reconstruction module is introduced to recover the lost roadside queries caused by communication interruptions, ensuring robustness.
    }
    \label{fig3}
\end{figure*}

\subsection{Cooperative Perception}
With the advancement of ITS, cooperative perception has rapidly progressed \cite{han2023collaborative,ruan2023learning,yu2024end}. First, multiple datasets \cite{yu2022dair,yu2023v2x,li2022v2x,xu2022opv2v,mao2022dolphins,xu2023v2v4real,zimmer2024tumtraf,ma2024holovic,hao2024rcooper} provide a foundation for subsequent research. In terms of methods, cooperation based on intermediate features achieves a trade-off between bandwidth and performance, thus becoming a mainstream approach \cite{han2023collaborative}.

To further reduce the transmission volume, several methods have been proposed, including convolution-based compression \cite{wang2020v2vnet,xu2022v2x}, selecting key regions \cite{hu2022where2comm}, and transmitting queries \cite{fan2023quest,chen2023transiff}. For fusion modules, many impressive works employ various methods, such as convolution neural networks \cite{wang2020v2vnet}, graph neural networks \cite{zhou2022multi}, knowledge distillation \cite{li2021learning}, and transformers \cite{lin2024v2vformer}.

However, previous methods only transmit non-temporal features and perform spatial fusion. Recently, some approaches have recognized the importance of temporal information. Some methods introduce temporal information at the transmitter to address the latency issue. FFNet \cite{yu2024flow} extracts feature flow from adjacent frames and sends it with the present features to EV for predicting BEV features. Similarly, RAO \cite{zhang2023robust} and \cite{dao2024practical} utilize consecutive point clouds for scene flow. How2comm \cite{yang2024how2comm} extracts feature flow from consecutive intermediate representations. CoBEVFlow \cite{wei2024asynchrony} records motion attributes and predicts flow using multi-head attention. On the other hand, some methods integrate temporal information into the fused features at EV. SCOPE \cite{yang2023spatio} fuses local historical features with current ones before aggregating multiple sources of information. Select2Col \cite{liu2024select2col} performs a weighted sum of EV's historical and fused features. UMC \cite{wang2023umc} and What2comm \cite{yang2023what2comm} utilize GRU-like components for temporal aggregation. HYDRO-3D \cite{meng2023hydro} and PragComm \cite{hu2024pragmatic} introduce a tracker following the cooperative detector to exploit the temporal features in trajectories. 

These methods, although accounting for temporal information, fail to fully leverage it to enhance the performance. Moreover, these temporal methods typically take point clouds or multi-modal data as inputs. Therefore, to fill the gap, we propose a camera-based temporal VIC3D object detection framework with multi-level temporal contexts integration, achieving maximal temporal benefits.

\section{METHODOLOGY}
In this section, we first briefly illustrate the pipeline of our CTCE. Subsequently, we provide the details of the multi-level temporal contexts integration mechanism, the motion-aware reconstruction module, and the training strategy.

\subsection{Overview of the Proposed Framework}
The VIC3D object detection system comprises two subsystems, ego-vehicle, and infrastructure, denoted as $\mathrm{V}$ and $\mathrm{I}$ respectively. In the infrastructure system, a backbone $f_{\mathrm{I\_backbone}}$ extracts features $F_\mathrm{I}^{(t)}$ from the input image $X_\mathrm{I}^{(t)}$ and a query generator $f_{\mathrm{I\_generator}}$ outputs object queries $Q_\mathrm{I}^{(t)}$:
\begin{align}
    F_\mathrm{I}^{(t)} = f_{\mathrm{I\_backbone}}(X_\mathrm{I}^{(t)}),\\
    Q_\mathrm{I}^{(t)} = f_{\mathrm{I\_generator}}(F_\mathrm{I}^{(t)}).
\end{align}
Here each query is represented by a $m$-dimensional embedding along with a reference point, which represents the object at both the instance and intermediate feature levels. Additionally, we retain the queries $\{Q_\mathrm{I}^{(t-1)}, Q_\mathrm{I}^{(t-2)}, \dots, Q_\mathrm{I}^{(t-k_1)}\}$ generated from the previous $k_1$ time steps.

As the initial step of our proposed MTCI mechanism, we employ the TCA module, denoted as $f_{\mathrm{TCA}}$, to integrate the queries $Q_\mathrm{I}^{(t)}$ with the past queries $\{Q_{\mathrm{I}}^{(t-\tau)}\}_{\tau=1}^{k_1}$, yielding the context-aware queries $\mathcal{Q}_\mathrm{I}^{(t)}$, as shown in the following:
\begin{align}
    \mathcal{Q}_\mathrm{I}^{(t)} = f_{\mathrm{TCA}}(Q_{\mathrm{I}}^{(t)}, \{Q_{\mathrm{I}}^{(t-\tau)}\}_{\tau=1}^{k_1}).
\end{align}
This temporal interaction enhances the query representation capability, leading to significant improvements in detection results.
RSU then transmits the top $m$ queries with the highest confidence via V2X communication. The EV receives the queries and stores $k_2$ frames roadside queries.

In the ego-vehicle system, we first employ the same approach as the roadside to obtain object queries $Q_\mathrm{V}^{(t)}$.
Subsequently, we utilize the TGF module $f_{\mathrm{TGF}}$ to generate the final cooperative queries $\mathcal{Q}_{\mathrm{C}}^{(t)}$. Initially, the ego queries $Q_{\mathrm{V}}^{(t)}$ are fused with the received queries $\mathcal{Q}_\mathrm{I}^{(t)}$. Then, guided by the stored roadside historical sequence $\{\mathcal{Q}_\mathrm{I}^{(t-\tau)}\}_{\tau=1}^{k_2}$, the queries further learn temporal features. At this point, MTCI is completed. Finally, we input the fused cooperative queries into the detection head, denoted as $f_{\mathrm{head}}$, to produce the detection outputs $\tilde{\mathcal{Y}}_{\mathrm{C}}^{(t)}$:
\begin{align}
    \mathcal{Q}_{\mathrm{C}}^{(t)} & = f_{\mathrm{TGF}}(Q_{\mathrm{V}}^{(t)}, \mathcal{Q}_\mathrm{I}^{(t)},
    \{\mathcal{Q}_\mathrm{I}^{(t-\tau)}\}_{\tau=1}^{k_2}), \\
    \tilde{\mathcal{Y}}_{\mathrm{C}}^{(t)} & = f_{\mathrm{head}}(\mathcal{Q}_{\mathrm{C}}^{(t)}).
\end{align}

We also consider communication interruptions in practical applications. To address this issue, the MAR module, denoted as $f_\mathrm{MAR}$, predicts the lost roadside queries $\mathcal{Q}_\mathrm{I}^{(t)}$ from the roadside historical sequence $\{\mathcal{Q}_\mathrm{I}^{(t-\tau)}\}_{\tau=1}^{k_2}$, defined as,
\begin{align}
    \Tilde{\mathcal{Q}}_\mathrm{I}^{(t)} = f_\mathrm{MAR}(
    \{\mathcal{Q}_\mathrm{I}^{(t-\tau)}\}_{\tau=1}^{k_2}),
\end{align}
where $\Tilde{\mathcal{Q}}_\mathrm{I}^{(t)}$ is the predicted queries. If communication interruption occurs, the system can still perform spatial-temporal cooperation with the predicted queries.

\subsection{Multi-Level Temporal Contexts Integration}
As depicted in Fig. \ref{fig3}, to efficiently exploit temporal information and enhance performance, we propose a multi-level interaction with the roadside historical sequence. Different previous methods \cite{yang2023spatio,liu2024select2col}, the rationale behind choosing the roadside sequence over the ego-vehicle or fused sequence lies in the elevated and stationary installation of RSU, which inherently imparts greater spatial-temporal coherence to its sequences compared to those derived from the ego-vehicle or fused sequences, both of which are confined to the ego-vehicle's perspective. 
Specifically, the TCA in RSU and TGF in ego-vehicle collectively achieve multi-level temporal contexts integration. The RSU stores queries generated by its query generator, while the ego-vehicle stores temporal queries received from the roadside. At each timestamp, they update sequences following the First-In-First-Out principle. Next, we will elaborate on TCA and TGF separately.

\subsubsection{Temporal Contexts Aggregation}
We first introduce temporal information at the roadside to enhance the modeling capability of queries. Motivated by StreamPETR \cite{wang2023exploring}, at each timestamp, the query generator of the RSU models potential objects from the current observations. These non-temporal queries are then fed into the TCA module, where they perform multi-head cross-attention with the queries from the past $k_1$ frames. 
Note that the data volume of temporal queries $\mathcal{Q}_\mathrm{I}^{(t)}$ is equivalent to that of non-temporal queries $Q_{\mathrm{I}}^{(t)}$, hence this process does not increase the transmission load.

\subsubsection{Temporal-Guided Fusion}

\begin{figure}[t]
    \centering
    \includegraphics[width=0.9\linewidth]{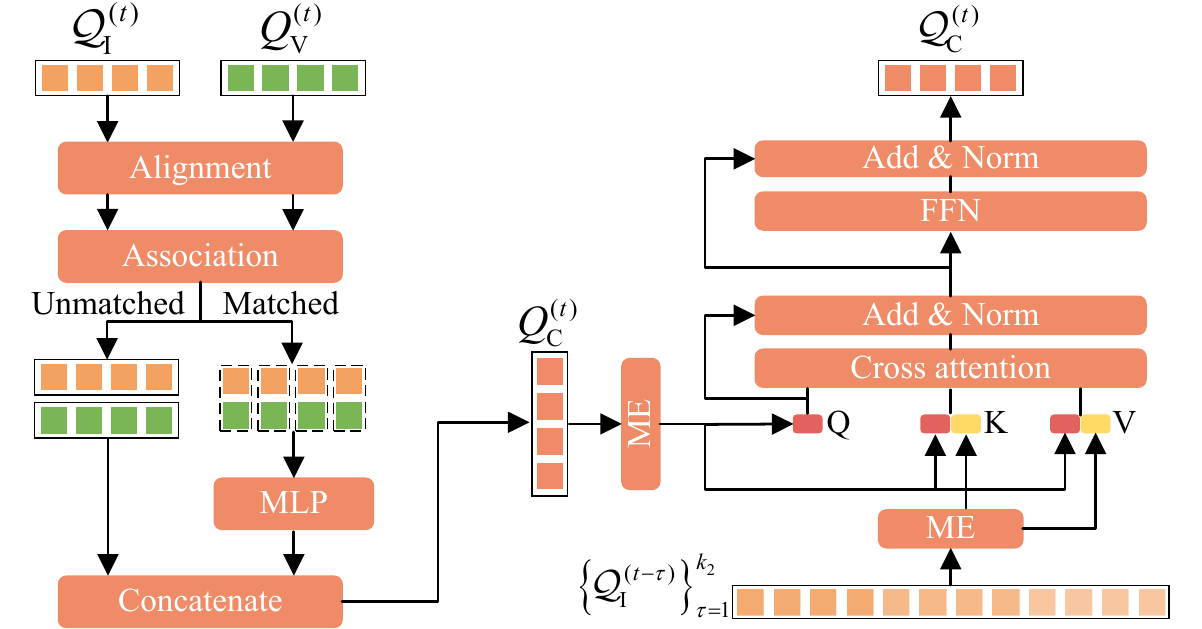}
    \vspace{-0.2cm}
    \caption{\textbf{Temporal-Guided Fusion Module.} This module first fuses roadside and ego-vehicle queries, and then the coarse fused queries interact with roadside historical queries to refine the representation by temporal contexts.
    }
    \label{fig4}
\end{figure}

Upon receiving the roadside queries, spatial-temporal fusion is further performed to propagate temporal information, as illustrated in Fig. \ref{fig4}.

At each timestamp, EV performs spatial fusion first. Due to the diverse perspectives of EV and RSU, a domain gap exists between the two sets of queries in both geometric and feature spaces. Domain alignment is required before fusion.

For the geometric space, the reference points are transformed into the unified coordinate system which is the EV's LiDAR coordinate system. For the feature space, position embeddings are obtained from reference points and then they are concatenated with the query embeddings to obtain the unified embeddings by an MLP.

Subsequently, we undertake the aggregation of the two distinct sets of queries. Given the shared observation regions between EV and RSU, it is necessary to identify instances concurrently detected by both agents. We construct a cost matrix based on the Euclidean distance between reference points and employ the Hungarian algorithm to obtain the best matches. For each pair of queries, embeddings are concatenated along the channel dimension, followed by fusion using an MLP. Queries that are not successfully matched are directly inserted into the coarse fused query set $Q_\mathrm{C}^{(t)}$.

To prevent the fused features from forgetting the contexts introduced by roadside queries, we further leverage the historical sequence from the roadside to guide the coarse fused queries to learn temporal features. Considering both EV and objects are moving, inspired by StreamPETR \cite{wang2023exploring}, a motion encoding (ME) module is introduced to implicitly encode timestamps and ego poses into the queries. Next, all $k_2$ frames of historical queries are concatenated as the \textit{key} and \textit{value}, while the coarse fused features serve as the \textit{query}. Temporal interactions are performed through multi-head cross-attention. It's worth noting that since the number of historical sequences is small ($k_2 \times m$), the temporal guidance refinement only incurs minimal computational overhead.

\subsection{Motion-Aware Reconstruction}
In the event of communication interruption, the EV is deprived of the additional information provided by RSU, resulting in the inability to benefit from cooperative perception. To alleviate the negative impact, we extract motion features from the roadside historical query sequence frame by frame to derive query trajectories, which are then employed to predict the queries that should have been received. 

\begin{figure}[t]
    \centering
    \includegraphics[width=0.9\linewidth]{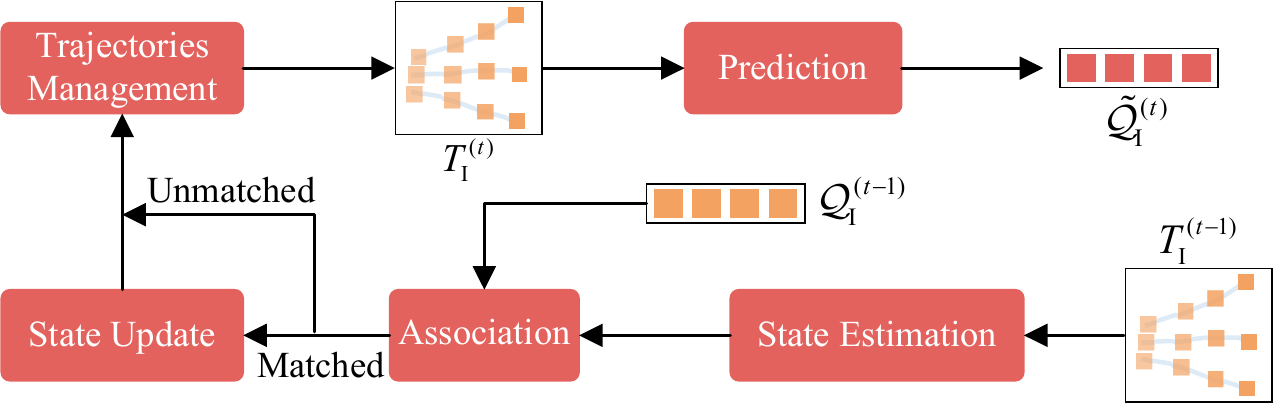}
    \vspace{-0.2cm}
    \caption{\textbf{The Pipeline of the Motion-Aware Reconstruction.} It first tracks queries from historical frames and then predicts the lost queries. This simple yet effective module helps gain robustness to communication interruptions.}
    \label{fig5}
\end{figure}

Specifically, we extract query trajectories based on the reference points and predict the reference points and query embeddings. For clarity in exposition, let us consider the prediction of queries at timestamp $t$, as depicted in Fig. \ref{fig5}. 

At timestamp $t-1$, to correlate the current queries $\mathcal{Q}_\mathrm{I}^{(t-1)}$ with historical trajectories $T_\mathrm{I}^{(t-1)}$, four essential procedures are required: i) state estimation, where a Kalman filter is used to estimate the states of historical trajectories $\Tilde{T}_\mathrm{I}^{(t-1)}$. A constant velocity model is employed during this process. Each trajectory state is described using a 6-dimensional vector $[x, y, z, v_x, v_y, v_z]^\mathrm{T}$, where $x$, $y$, $z$ denote the spatial coordinates and $v_x$, $v_y$ and $v_z$ indicate the velocities along the respective axes; ii) association, which constructs an affinity matrix based on the Euclidean distance between the queries $\mathcal{Q}_\mathrm{I}^{(t-1)}$ and estimated states $\Tilde{T}_\mathrm{I}^{(t-1)}$. Subsequently, the Hungarian algorithm is applied to solve the bipartite matching problem.
iii) state update, where the Kalman filter updates the matched trajectories based on their paired reference points; iv) trajectory management, which creates new trajectories for newly appeared instances and removes disappeared ones to obtain the final trajectory results $T_\mathrm{I}^{(t)}$. It is noteworthy that despite tracking queries based on reference points, we preserve the embedding of each query. Consequently, we refer to the outcome as query trajectories.

At timestamp $t$, EV fails to receive the roadside queries because of the interruptions, so the reconstruction process is initiated. Regarding the reference points, we directly ascertain their current positions by leveraging the states of the trajectories. Since the instances in traffic scenes exhibit continuous motion trends over a short period \cite{scholler2020constant}, we employ a constant velocity model for prediction.
Due to the powerful sequential modeling capability, we employ multi-head cross-attention to forecast the query embeddings. First, historical timestamps are encoded into time embeddings using cosine and sine functions. These embeddings are then added to the corresponding historical query embeddings, and the resulting features are concatenated along the time, serving as the \textit{key} and \textit{value}. Additionally, the current timestamp is encoded in the same manner as the \textit{query}.

Finally, by combining the predicted reference points and query embeddings, we recover the lost roadside queries $\Tilde{\mathcal{Q}}_\mathrm{I}^{(t)}$ in a simple yet effective manner. Then EV continues with the spatial-temporal cooperation as per standard procedure, akin to normal conditions.

\subsection{Training Strategy and Loss Functions}

To ensure strong cooperative perception capabilities in the framework, even under communication interruptions, we propose a two-step training strategy.

Initially, we train the framework without the MAR module using the corresponding cooperative ground truth $\mathcal{Y}_{\mathrm{C}}^{(t)}$ under ideal conditions. During this phase, we utilize focal loss $L_{\mathrm{focal}}$ for classification and smooth L1 loss $L_{\mathrm{L1}}$ for regression to provide supervision to the network. The final loss is defined as the weighted sum of both, as follows:
\begin{align}
    \mathcal{L}_{\mathrm{stage1}} =
    \alpha L_{\mathrm{focal}}( \Tilde{\mathcal{Y}}_{\mathrm{C}}^{(t)}, \mathcal{Y}_{\mathrm{C}}^{(t)}) + \beta L_{\mathrm{L1}}(\Tilde{\mathcal{Y}}_{\mathrm{C}}^{(t)}, \mathcal{Y}_{\mathrm{C}}^{(t)}),
\end{align}
where $\alpha$ and $\beta$ are coefficients.

Building upon the first step, we train the MAR module in a self-supervised manner. Specifically, we take continuous $k_2+1$ frames of roadside data as a training sample. Using the first $k_2$ frames, we predict the queries for the last frame and compute the mean squared loss $L_{\mathrm{MSE}}$ between the prediction and the ground truth, which can be formulated as follows:
\begin{align}
    \mathcal{L}_{\mathrm{stage2}} = L_{\mathrm{MSE}}(\Tilde{\mathcal{Q}}_\mathrm{I}^{(t)}, \mathcal{Q}_\mathrm{I}^{(t)}).
\end{align}

It is notable that by adopting this training methodology, the resulting framework demonstrates adaptability to fluctuations in communication conditions without specialized training for specific PDRs.

\section{Experiments}

\subsection{Experimental Setup}
\subsubsection{Datasets} 
We conduct experiments on two datasets, and the details are as follows:

\textbf{V2X-Seq dataset} \cite{yu2023v2x} is tailored for vehicle-infrastructure cooperative perception collected from the real world, comprising over 15,000 frames captured at a frequency of 10 Hz across 95 scenes. Both EV and RSU are equipped with a monocular camera. We conduct comparative experiments and ablation studies on this dataset, reporting the results on the validation set. The perception range of EV is set to $x\in[-51.2, 51.2]$ and $y\in[-51.2,51.2]$.

\textbf{V2X-Sim dataset} \cite{li2022v2x} is collected from simulated scenes in Carla. It comprises 100 sequences, each containing 100 frames of data. To validate the generalizability of our approach, we conduct vehicle-to-vehicle (V2V) cooperative perception experiments on this dataset. For simplicity and efficiency during training, we select two CAVs in each scene, each equipped with three forward-facing cameras. The detection range is set to $[-32.0, -32.0, 32.0, 32.0]$.

\subsubsection{Evaluation Metrics}
We adopt several metrics to evaluate the performance, including mean Average Precision (mAP) and three True Positive (TP) metrics: mean Average Translation Error (mATE), mean Average Scale Error (mASE), and mean Average Orientation Error (mAOE) \cite{caesar2020nuscenes}. Given its comprehensive reflection of performance, we prioritize mAP as the primary metric.

\subsection{Implementation Details}
We select individual perception \cite{li2022bevformer} without cooperation and Late Fusion \cite{xu2022opv2v} as two baselines and compare our method with the state-of-the-art (SOTA) cooperative methods, including V2VNet \cite{wang2020v2vnet}, V2X-ViT \cite{xu2022v2x}, Where2comm \cite{hu2022where2comm}, QUEST \cite{fan2023quest} and DiscoNet \cite{li2021learning}.

For a fair comparison we adopt BEVFormer \cite{li2022bevformer} as the foundation model for all methods, employing a ResNet101 pretrained on FCOS3D as the image backbone. We use the AdamW optimizer with a weight decay of $0.01$ to train the models of the proposed framework. The initial learning rate is set to $2\times10^{-4}$, and cosine annealing is applied for the schedule. For the first training stage, we set $\alpha=2.0$ and $\beta=0.25$. For temporal modeling, we set $k_1=4$, $k_2=4$ and $m=256$.

\subsection{Quantitative Results}
\subsubsection{Performance Comparison} 
Table \ref{table1} reports the comparative detection performance of different methods on the V2X-Seq and V2X-Sim datasets, and CTCE shows superior performance on mAP. Firstly, in comparison to individual perception, CTCE achieves a notable enhancement of $34.9\%$ in mAP, coupled with reductions of 0.202 and 0.126 in mATE and mAOE, respectively. Furthermore, our approach outperforms state-of-the-art (SOTA) methods, exhibiting a $3.8\%$ increase in mAP compared to QUEST on V2X-Seq. Additionally, we observe a reduction of 0.054 in mATE, suggesting that leveraging multi-level temporal information enhances accuracy and localization. 
For V2V cooperative detection, CTCE improves mAP by $1.3\%$ compared to QUEST, which demonstrates the effectiveness of our method in V2V scenarios. This slight improvement may be attributed to the difficulty of integrating temporal information from the vehicle-side historical sequence due to occlusion and motion. 

\subsubsection{Robustness to Communication Interruptions} We verify the robustness to communication interruptions of our proposed framework on the V2X-Seq dataset. Fig. \ref{fig6} illustrates the performance of our method and other cooperative methods under varying PDRs. We observe that our proposed framework outperforms others across all settings. Moreover, as the PDR increases from $0$ to $50\%$, our method only experiences a $3.4\%$ decrease in mAP. Even under extreme communication conditions with a PDR of $80\%$, our method still outperforms the Individual perception by a large margin, achieving approximately twice the mAP. This proves that our proposed MAR can assist the framework in robustness against communication interruptions, offering promising prospects for practical applications.

\begin{table}[tbp]
    \setlength{\tabcolsep}{0.9pt}
    \caption{Detection Performance on V2X-Seq and V2X-Sim Datasets.}
    \label{table1}
    \begin{tabular*}{1.08\columnwidth}{ c | c c c c | c c c c}
        \hline\hline
        \multirow{2}{*}{\centering Model} & \multicolumn{4}{c|}{V2X-Seq} & \multicolumn{4}{c}{V2X-Sim} \\
        \centering ~ & mAP \hspace{-0.75em} $\uparrow$ & mATE \hspace{-0.75em} $\downarrow$ & mASE \hspace{-0.75em} $\downarrow$ & mAOE \hspace{-0.75em} $\downarrow$ & mAP \hspace{-0.75em} $\uparrow$ & mATE \hspace{-0.75em} $\downarrow$ & mASE \hspace{-0.75em} $\downarrow$ & mAOE \hspace{-0.75em} $\downarrow$ \\
        \hline
        \centering Individual \cite{li2022bevformer} & 0.155 & 0.825 & 0.144 & 0.279 & 0.161 & 0.811 & 0.232 & 0.301\\
        \centering Late Fusion \cite{xu2022opv2v}& 0.419 & 0.627 & 0.148 & 0.125 & 0.198 & 0.808 & 0.222 & 0.267\\
        \centering V2VNet \cite{wang2020v2vnet} & 0.430 & 0.631 & 0.158 & 0.221 & 0.214 & 0.768 & 0.255 & 0.349 \\
        \centering DiscoNet \cite{li2021learning} & 0.358 & 0.622 & 0.162 & 0.190 & 0.220 & 0.787 & 0.267 & 0.411\\
        \centering V2X-ViT \cite{xu2022v2x} & 0.352 & 0.652 & 0.166 & 0.186 & 0.224 & 0.848 & 0.250 & 0.383\\
        \centering Where2comm \cite{hu2022where2comm} & 0.350 & 0.746 & 0.161 & 0.383 & 0.190 & 0.911 & 0.275 & 0.310\\
        \centering QUEST \cite{fan2023quest}& 0.466 & 0.677 & 0.158 & 0.113 & 0.239 & 0.832 & 0.259 & 0.390\\
        \hline
        \centering \textbf{CTCE(Ours)} & \textbf{0.504} & 0.623 & 0.158 & 0.153 & \textbf{0.252} & 0.843 & 0.259 & 0.324 \\
        \hline\hline
    \end{tabular*}
\end{table}

\subsection{Qualitative Results}
As shown in Fig. \ref{fig7}, we present the visualized detection results on the V2X-Seq dataset, providing an intuitive representation of detection performance. We provide the results of QUEST and CTCE under both ideal communication and communication interruption.

It can be observed that compared to QUEST, our method achieves more accurate localization with the enhancement of temporal contexts. When communication interruptions occur, our method can recover the lost information and still achieve cooperative perception. In contrast, QUEST fails to detect blind-spot objects, degrading to Individual perception.

\subsection{Ablation Studies and Analysis}

\subsubsection{Importance of Multi-Level Temporal Contexts Integration} As shown in Table \ref{table2}, we progressively add i) TCA, ii) TGF, and iii) the number of frames in TGF, train the models, and evaluate their performance on V2X-Seq. The results indicate that both TCA and TGF enhance performance through temporal information, with mAP improving by $1.8\%$ and $0.9\%$, respectively. Furthermore, combining them, i.e., multi-level temporal information integration, yields the maximum benefit from temporal information, resulting in a $3.8\%$ mAP improvement. For the effect of the number of previous frames $k_2$, the mAP improves as it increases, reaching saturation at 4 frames. That's because, with the assistance of TCA, TGA can establish long-term dependencies without interacting with very long historical sequences.

\begin{figure}[t]
    \centering
    \includegraphics[width=0.85\linewidth]{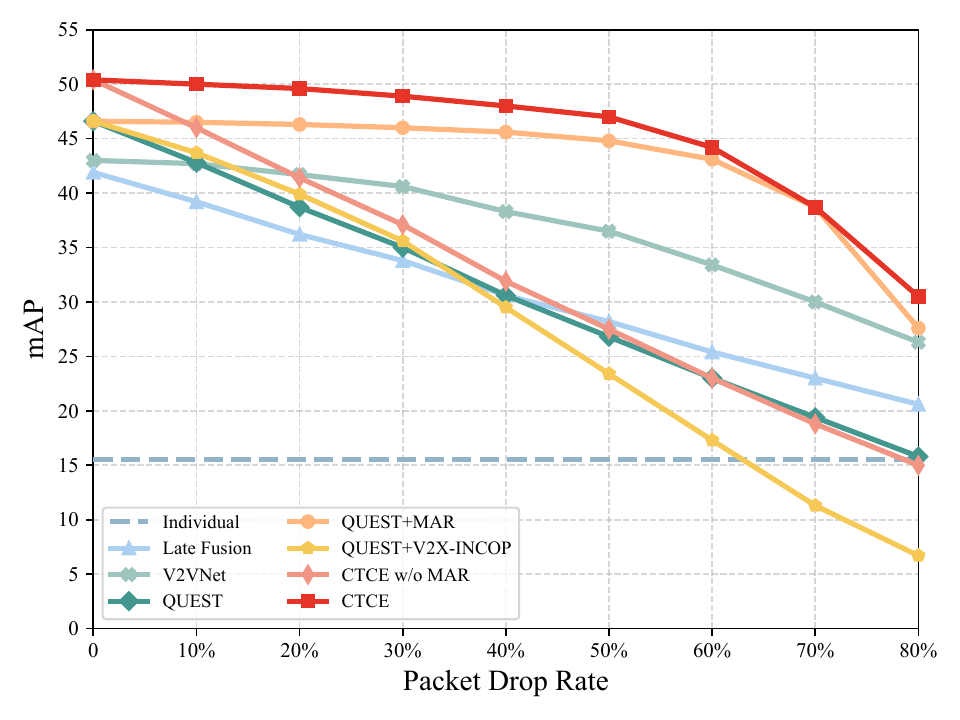}
    \vspace{-0.2cm}
    \caption{\textbf{Detection Performance under Different Packet Drop Rates.} Thanks to the proposed MAR, CTCE outperforms all other methods across all packet drop rate settings and exhibits robustness to communication interruptions.}
    \label{fig6}
\end{figure}

\begin{figure*}[t]
    \centering
    \subfloat[Images of the First Scene in V2X-Seq]{
        \begin{minipage}[b]{0.42\linewidth}
            \includegraphics[width=0.5\linewidth]{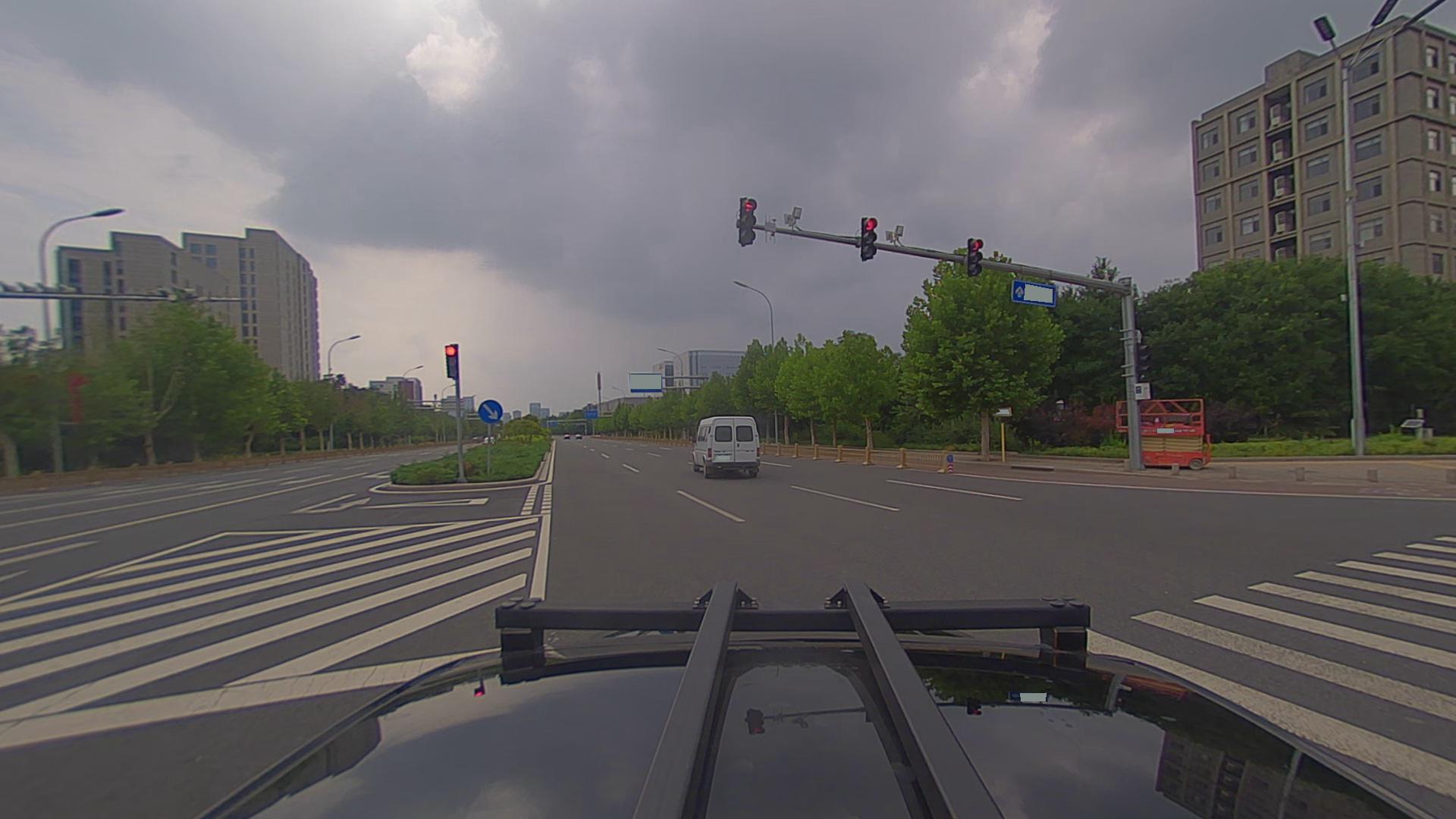}%
            \includegraphics[width=0.5\linewidth]{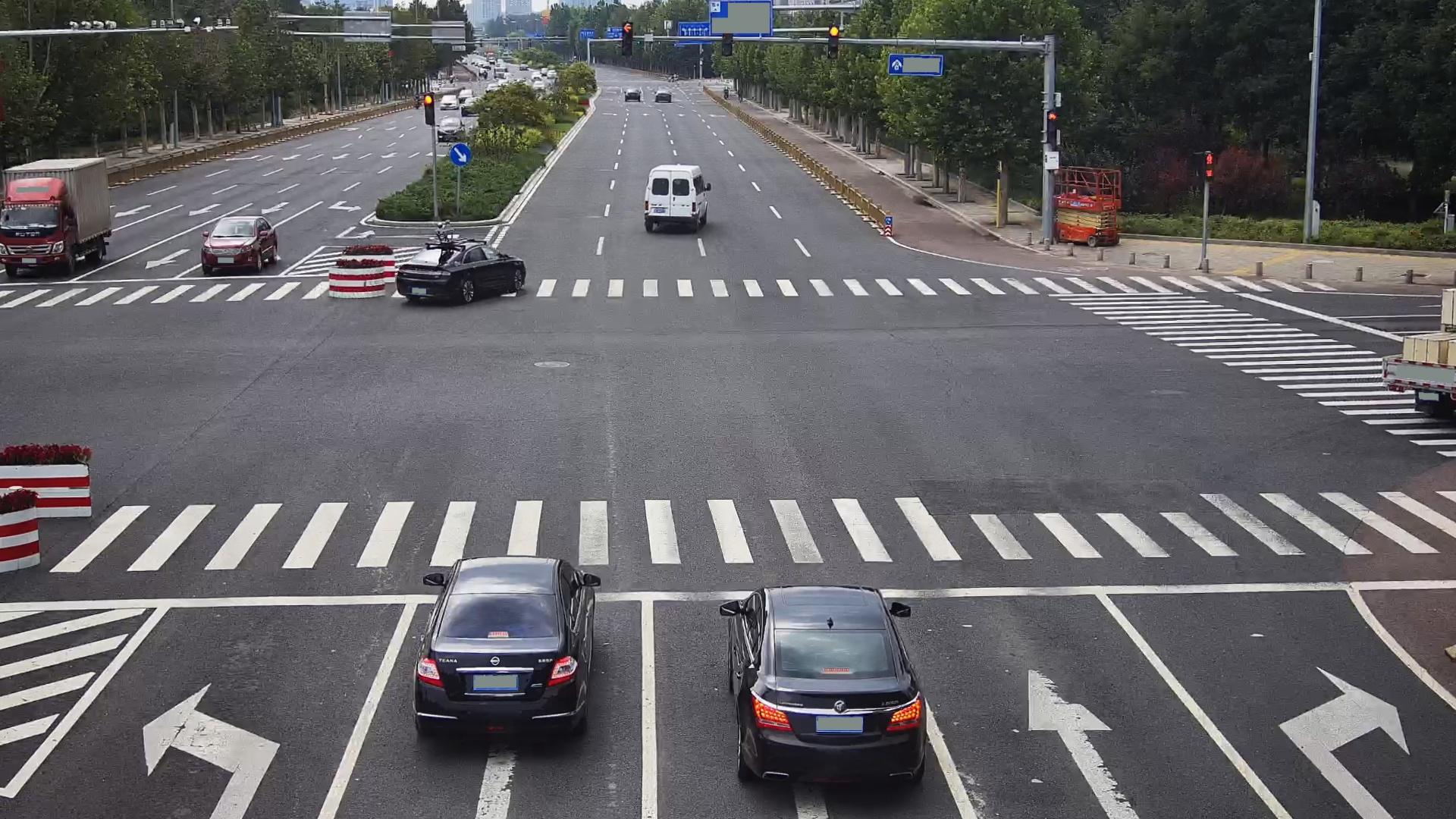}%
        \end{minipage}
        \label{fig7a}
    }%
    \subfloat[Images of the Second Scene in V2X-Seq]{
        \begin{minipage}[b]{0.42\linewidth}
            \includegraphics[width=0.5\linewidth]{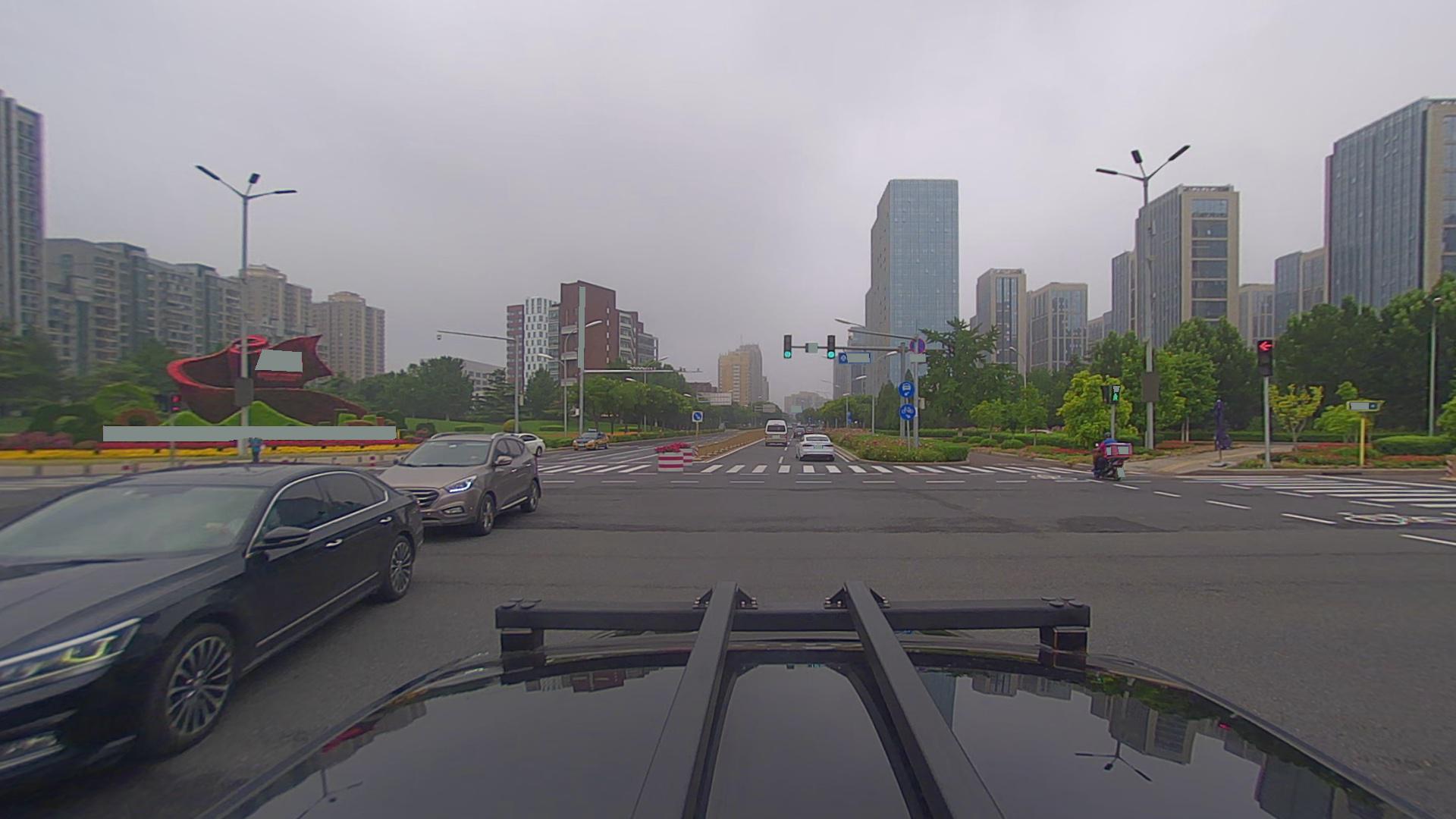}%
            \includegraphics[width=0.5\linewidth]{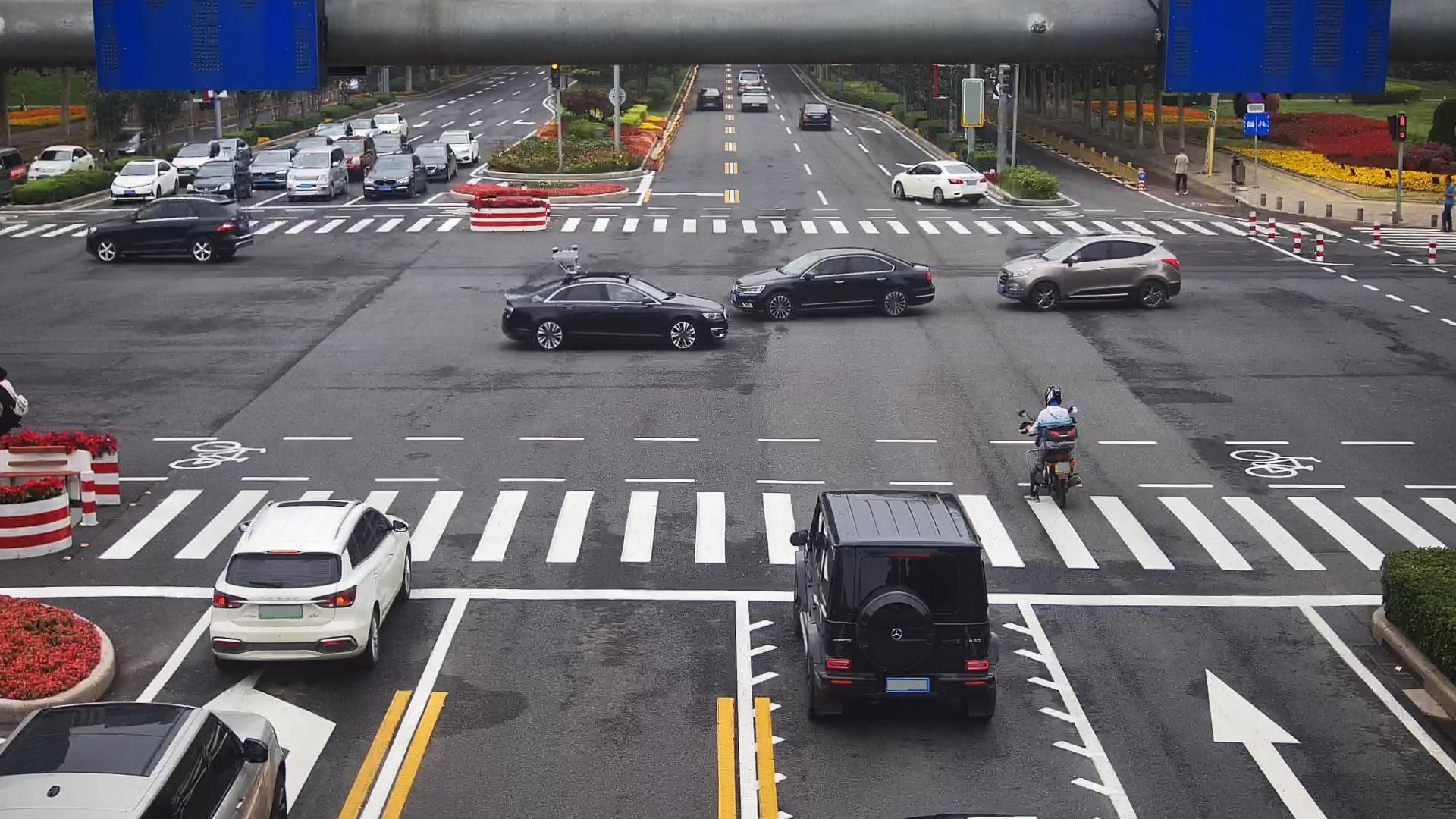}%
        \end{minipage}
        \label{fig7b}
    }%
    \vspace{-0.7\baselineskip}
    \\
    \centering
    \subfloat[CTCE]{
        \begin{minipage}[b]{0.21\linewidth}
        \centering
            \includegraphics[width=1.0\linewidth]{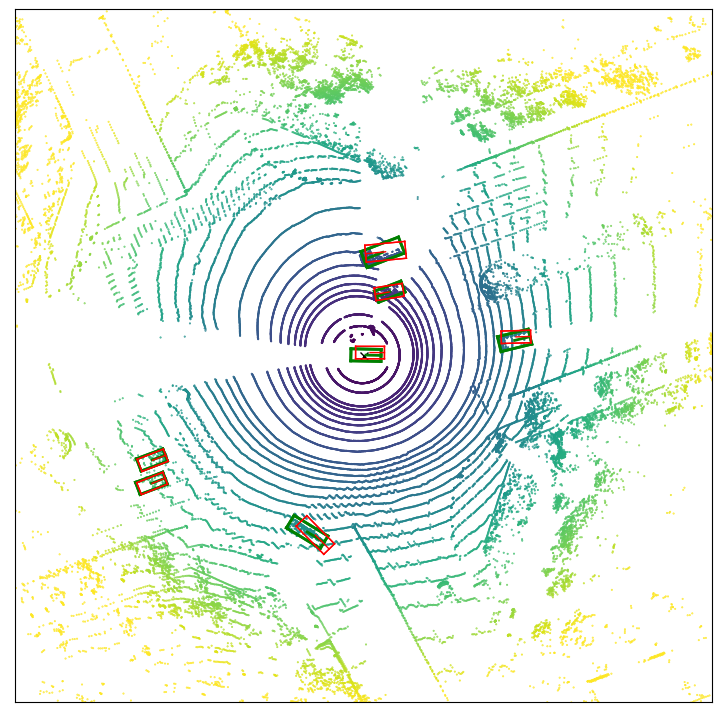}\\
            \includegraphics[width=1.0\linewidth]{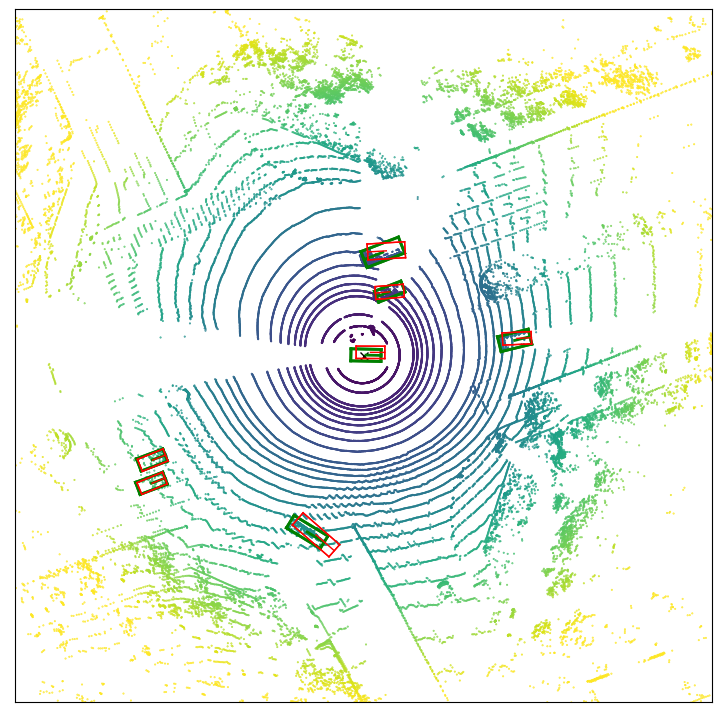}
        \end{minipage}
        \label{fig7c}
    }
    \centering
    \subfloat[QUEST]{
        \begin{minipage}[b]{0.21\linewidth}
        \centering
            \includegraphics[width=1.0\linewidth]{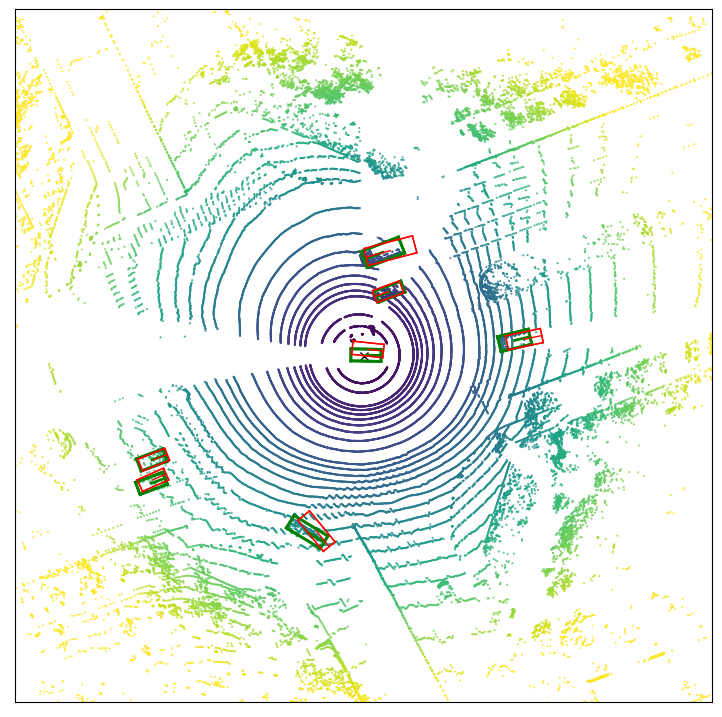} \\
            \includegraphics[width=1.0\linewidth]{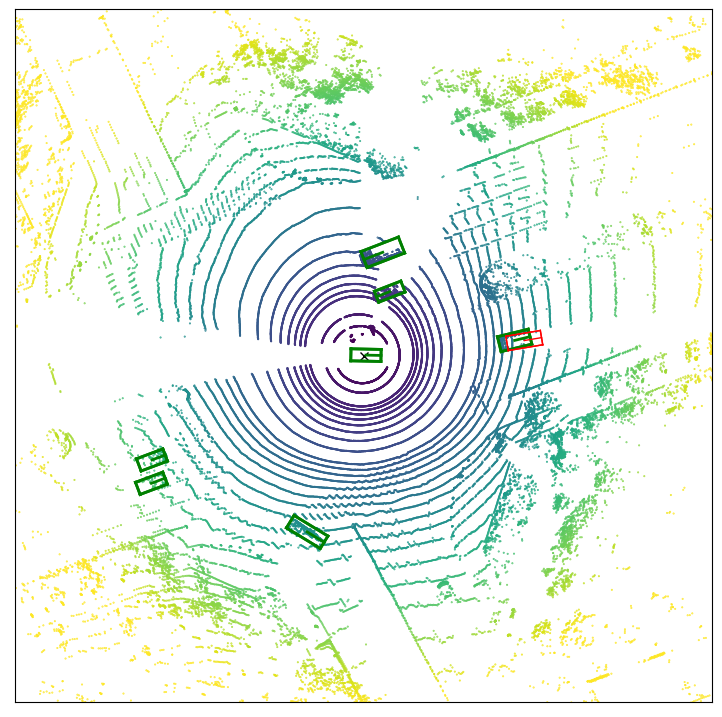}
        \end{minipage}
        \label{fig7d}
    }
    \centering
    \subfloat[CTCE]{
        \begin{minipage}[b]{0.21\linewidth}
        \centering
            \includegraphics[width=1.0\linewidth]{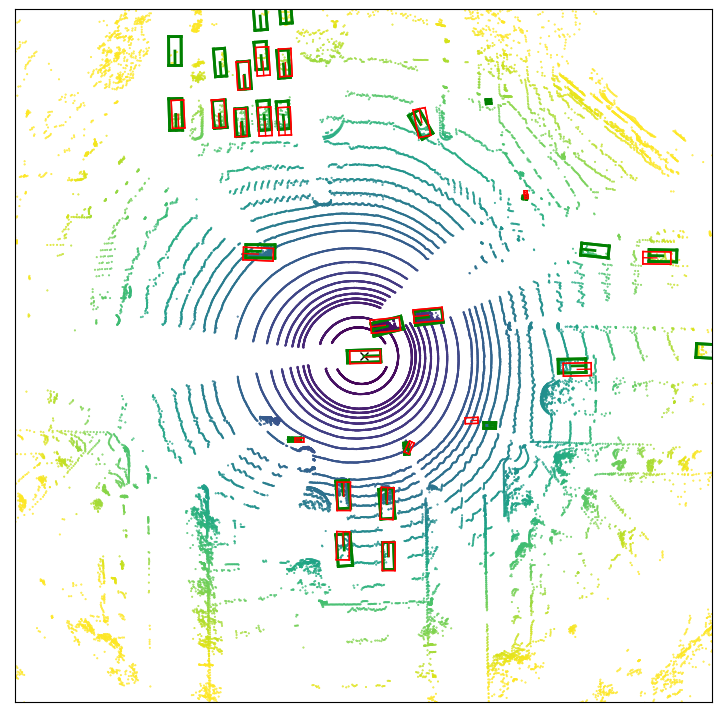} \\
            \includegraphics[width=1.0\linewidth]{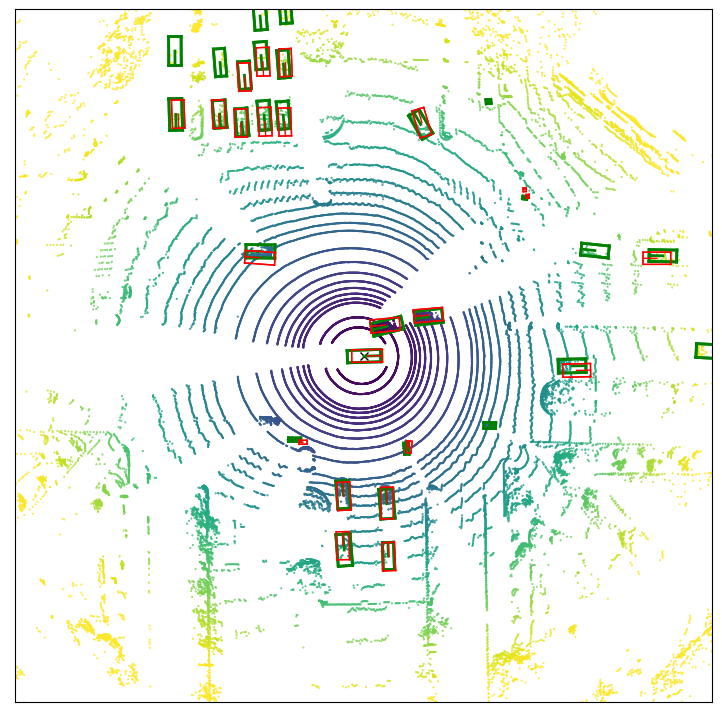}
        \end{minipage}
        \label{fig7e}
    }
    \centering
    \subfloat[QUEST]{
        \begin{minipage}[b]{0.21\linewidth}
        \centering
            \includegraphics[width=1.0\linewidth]{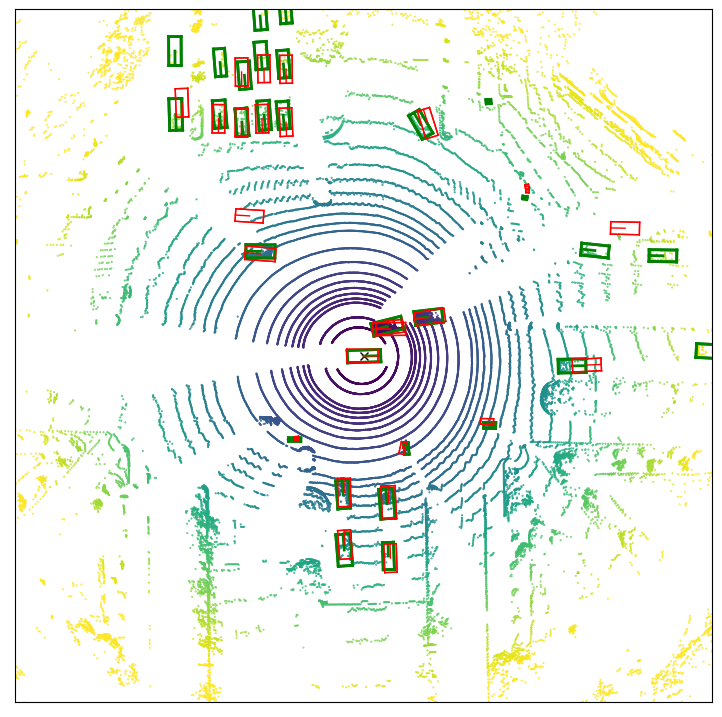} \\
            \includegraphics[width=1.0\linewidth]{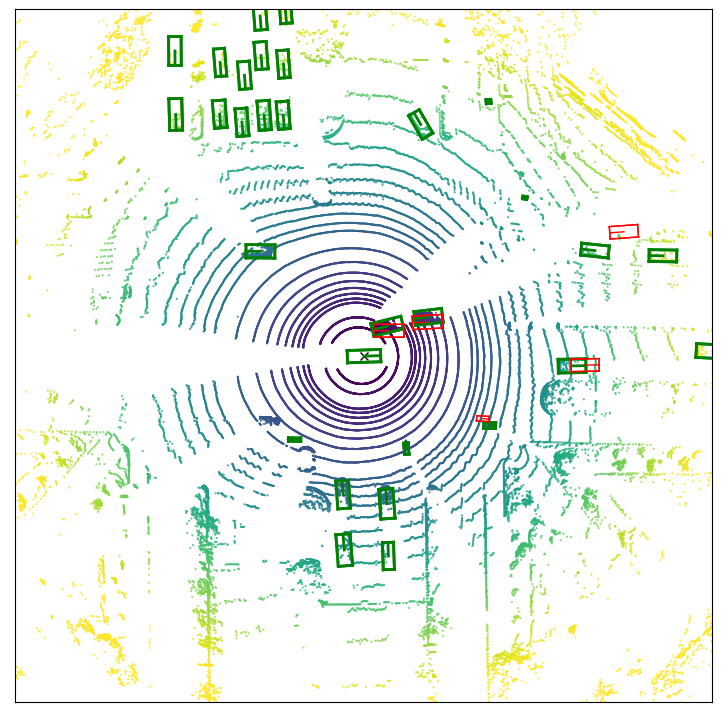}
        \end{minipage}
        \label{fig7f}
    }
    \caption{\textbf{Visualization with V2X-Seq.} \protect\subref{fig7a} and \protect\subref{fig7b} present raw images from two typical scenarios in V2X-Seq. \protect\subref{fig7c}-\protect\subref{fig7f} display detection results under both ideal communication conditions and communication interruption using our CTCE and QUEST methods. The upper row shows results under ideal communication conditions, while the lower row illustrates results during communication interruptions. Ground truth is represented by green rectangles, and predictions are shown in red rectangles.}
    \label{fig7}
\end{figure*}

\begin{table}[tbp]
    \caption{Importance of Multi-Level Temporal Contexts Integration}
    \label{table2}
    \centering
    \begin{tabular*}{0.9\columnwidth}{@{\extracolsep{\fill}} c c c | c c c c}
        \hline\hline
        TCA & TGF & $k_2$ & mAP & mATE & mASE & mAOE \\
        \hline
        ~ & ~ & ~ & 0.466 & 0.677 & 0.158 & 0.113 \\
        ~ & \checkmark & 4 & 0.475 & 0.623 & 0.156 & 0.108 \\
        \checkmark & ~ & ~ & 0.484 & 0.616 & 0.153 & 0.120\\
        \hline
        \checkmark & \checkmark & 1 & 0.478 & 0.639 & 0.166 & 0.153\\
        \checkmark & \checkmark & 2 & 0.492 & 0.637 & 0.151 & 0.153\\
        \checkmark & \checkmark & 3 & 0.497 & 0.622 & 0.158 & 0.152 \\
        \checkmark & \checkmark & 4 & \textbf{0.504} & 0.623 & 0.158 & 0.153 \\
        \checkmark & \checkmark & 5 & 0.498 & 0.625 & 0.163 & 0.150\\
        \hline\hline
    \end{tabular*}
\end{table}

\subsubsection{Role of MAR in Robustness to Interruptions}
As shown in Fig. \ref{fig6}, to investigate the effect of MAR, we design 3 models. First, we remove the MAR from the CTCE, termed as \textit{CTCE w/o MAR}. Compared with CTCE, the mAP of \textit{CTCE w/o MAR} and PDR are approximately linearly negatively correlated. Then we incorporate MAR into the QUEST, denoted as \textit{QUEST+MAR}, which also exhibits robustness to interruptions, with mAP decreasing by only $3.5\%$ when PDR is $50\%$. These comparisons demonstrate that MAR is effective against interruptions and can be used as a plugin for other query-based cooperative frameworks.

Moreover, an alternative solution is to recover the lost queries from the fused queries \cite{ren2023interruption}. Motivated by this idea, we implement \textit{QUEST+V2X-INCOP} based on QUEST. The results indicate that this method improves the performance of QUEST at lower PDRs. However, as communication conditions deteriorate, it negatively affects the original performance of QUEST. This discrepancy arises from the differing coverage areas of fused queries and roadside queries. It becomes challenging to fully restore lost roadside queries from fused queries, particularly under extreme communication conditions, thereby undermining overall performance. Therefore, we propose reconstructing lost queries from the historical roadside sequence.

\subsubsection{Research on the Temporal Integration Strategy of TGF}
TGF utilizes roadside historical sequence to enhance performance, while alternative strategies use historical ego features \cite{yang2023spatio} or fused features \cite{liu2024select2col}.
We re-implemented the TGF module according to the two strategies mentioned above, referred to as \textit{Ego Queries} and \textit{Fused Queries}. Additionally, we disabled the TGF, labeled as \textit{No Queries} serving as a baseline.
As shown in Table \ref{table3}, all strategies exhibit an improvement in mAP with temporal information. But \textit{Roadside Queries} demonstrate superior performance compared to the others. This is attributed to the stronger spatial-temporal consistency inherent in roadside historical sequences.

\begin{table}[!t]
    \caption{Ablation Study of Different Temporal Integration Strategy}
    \label{table3}
    \centering
    \begin{tabular*}{0.9\columnwidth}{@{\extracolsep{\fill}} p{0.25\columnwidth} | c c c c}
        \hline\hline
        \centering Strategy & mAP & mATE & mASE & mAOE \\
        \hline
        \centering No Queries & 0.484 & 0.616 & 0.153 & 0.120 \\
        \centering Ego Queries & 0.490 & 0.635 & 0.160 & 0.176 \\
        \centering Fused Queries & 0.495 & 0.650 & 0.154 & 0.132 \\
        \centering \textbf{Roadside Queries} & \textbf{0.504} & 0.623 & 0.158 & 0.153 \\
        \hline\hline
    \end{tabular*}
\end{table}

\section{CONCLUSIONS}
This paper introduces CTCE, the first camera-based temporal cooperative 3D object detection framework. Our method enhances performance by integrating temporal contexts at multiple levels and includes a motion-aware reconstruction module to recover lost queries during communication interruptions. Experimental results demonstrate CTCE's superiority over other methods and its robustness to interruptions. We hope that this work can advance the practical application of intelligent transportation systems. In the future, we will explore leveraging temporal contexts to address additional real-world challenges such as localization errors and communication delays.

\bibliographystyle{IEEEtran}
\bibliography{reference}

\end{document}